# Pavement Fatigue Crack Detection and Severity Classification Based on Convolutional Neural Network

Zhen Wang

Center for Smart and Green Civil Systems
The Charles E. Via, Jr. Department of Civil and Environmental Engineering
Virginia Polytechnic Institute and State University
750 Drillfield Drive
Blacksburg, VA, USA 24060
Email: wzhen@vt.edu
Phone: (540) 315-0302

Dylan G. Ildefonzo

Graduate Research Assistant
Center for Smart and Green Civil Systems
The Charles E. Via, Jr. Department of Civil and Environmental Engineering
Virginia Polytechnic Institute and State University
750 Drillfield Drive
Blacksburg, VA, USA 24060
Email: ildefonzo@vt.edu
Phone: (757) 816-1605

Linbing Wang

Professor
Center for Smart and Green Civil Systems
The Charles E. Via, Jr. Department of Civil and Environmental Engineering
Virginia Polytechnic Institute and State University
750 Drillfield Drive
Blacksburg, VA, USA 24060
Email: wangl@vt.edu
Phone: (540) 231-5262

**Abstract**

Due to the varying intensity of pavement cracks, the complexity of topological structure, and the noise of texture background, image classification for asphalt pavement cracking has proven to be a challenging problem. Fatigue cracking, also known as alligator cracking, is one of the common distresses of asphalt pavement. It is thus important to detect and monitor the condition of alligator cracking on roadway pavements. Most research in this area has typically focused on pixel-level detection of cracking using limited datasets. A novel deep convolutional neural network that can achieve two objectives is proposed. The first objective of the proposed neural network is to classify presence of fatigue cracking based on pavement surface images. The second objective is to classify the fatigue cracking severity level based on the Distress Identification Manual (DIM) standard. In this paper, a databank of 4484 high-resolution pavement surface images is established in which images are taken locally in the Town of Blacksburg, Virginia, USA. In the data pre-preparation, over 4000 images are labeled into 4 categories manually according to DIM standards. A four-layer convolutional neural network model is then built to achieve the goal of classification of images by pavement crack severity category. The trained model reached the highest accuracy among all existing methods. After only 30 epochs of training, the model achieved a crack existence classification accuracy of 96.23% and a severity level classification accuracy of 96.74%. After 20 epochs of training, the model achieved a pavement marking presence classification accuracy of 97.64%.

*Keywords*: fatigue cracking, alligator cracking, pavement marking, pavement severity, deep convolutional neural networks, supervised machine learning, crack detection, Tensorflow

## 1. Introduction

*1.1. Background*

The survey and analysis of pavement distresses is vital for pavement safety, maintenance, and evaluation [1,2]. Crack generation and severity level directly reflect the condition of the pavement. Therefore, accurate detection and quantification of pavement cracks with complex physical topologies can deliver sufficient information for quantification of pavement quality, and for prevention of pavement distresses [3]. The traditional manual inspection method is labor intensive, requires time-consuming manual data processing, and can be potentially hazardous to inspectors and road users alike [4]. As such, research on automated crack detection and classification has drawn great attention in recent years from industry professionals and academics alike.

The use of intelligent image processing algorithms for pavement crack identification is one example of automated pavement crack detection. The goal of these intelligent image detection algorithms is to use readily-accessible camera technology to detect pavement cracks and then classify pavement condition based on the presence of those cracks. Image binarization, mathematical morphology, and edge detection are widely used algorithms for image processing. In image binarization algorithms, image pixels are converted into one of two values: black or white. In photographs, cracked areas always appear darker than the non-cracked areas. When a binarization algorithm is applied to a photograph, cracked areas present as black pixels in the images, while non-cracked areas present as white pixels. Mathematical morphology can further be used to modify the binarized shapes to achieve better quality crack detection. Edge detection can also be used to create a distinct border between pixels representing cracked and non-cracked areas [5].

Crack detection methods have been suggested based on image processing. Generally, these methods fall into one of two categories: conventional methods and machine learning methods. Conventional methods attempt to find suitable thresholds to isolate cracks or apply edge detection algorithms to identify the cracks from input images. Conventional methods typically do not produce accurate detection results due to their inability to handle variation in crack shape and size, as well as their inability to overcome various sources of image noise, such as nonhomogeneous shading and color intensity. In order to overcome the shortcomings of conventional methods, modern machine learning (ML) based methods have been proposed.

In ML based methods, deep learning based (DL) approaches have achieved significant performance advantages in detection, classification, and segmentation problems without any assumption of data distribution. DL methods have proven to be useful tools for the automatic extraction of deep features from raw image data through the use of a convolutional neural network (CNN). These deep networks are a useful tool for learning and adjusting network parameters based on training data samples without the need for user-defined predefinition of features. However, millions of data points are typically necessary to train a good-quality model by this method [6,7].

The approach in this paper is inspired by a CNN model proposed for the ImageNet classification contest [8]. In the ImageNet classification contest, 1000 categories of images are classified. For pavement condition management, pavement fatigue crack detection based on image classification has many fewer categories.

An early attempt of use of an artificial neural network (ANN) to detect crack on images of Portland cement concrete structures was proposed in 2011 [9]. In this work, a backpropagation neural network is applied to automate image classification. The network was trained on 105 concrete images, after which point performance was validated using 120 new images. The accuracy of recognition of images containing cracks (crack images) and images containing no cracks (non-crack images) were 90% and 92%, respectively.

The early pavement distress image database German Asphalt Pavement Disease (GAP) has been established for deep learning applications. The dataset has 1,969 grayscale road images including cracks (longitudinal/lateral, crocodile, seal/fill) pits, patches, openings, seams and bleeding [10]. Smartphone and unmanned aerial vehicle (UAV) are applied as a powerful tool of image database building. Several training databases are established by high resolution smartphones [11-13].

Due to the morphological similarities between crack environment noise, many modified CNN models are presented to achieve better accuracy of crack image and crack pixel detection. ConvNet uses corrector linear unit (ReLU) activation functions (usually used with DCNNs) with dropout regularization method. Random gradient descent optimization was employed in the training process [14]. Principal component analysis (PCA) was adopted to classify different crack types including longitudinal, transverse and alligator cracks in the analysis image grid [15].

A machine learning model to detect cracks in concrete surface images by considering crack-like noise was proposed in 2017 based on Crack Candidate Region (CCR) to classify cracks and objects that share similar features with cracks [13]. The VGG-16 model was applied in their architecture which is a pretrained 16-layer DCNN by ImageNet. The CrackNet-V network was modified to better detect pixel-level crack on pavement images [16]. Also, a modified U-shape network for detecting and segmenting crack distresses from the input images was proposed. A fusion layer was applied to merge the features extracted from two branch networks. The architecture of the proposed network combined the modified U-network and High-level Feature network with a fusion layer.

Most approaches mentioned focus on crack pixel detection based on deep learning approach [13-19,21,23]. However, in Pavement Management System (PMS) or pavement condition rating system, detecting the existence of crack images in huge dataset is the primary goal. For rating systems or pavement maintenance, classifying different crack severity levels is more important. The existing research on crack image classification does not exhibit sufficient accuracy for these applications. Further, crack severity classification is not included in any of the current approaches [11,12,20,22,24].

*1.2. Significance of Work Presented*

In the research presented, a locally collected image dataset was used to train a classifier network of the CNN subtype. The images acquired were divided into thousands of training image blocks. The primary goal of the network created is to achieve reliable execution of fatigue crack existence classification and fatigue cracking severity level classification. No existing research has been successful in the classification of multi-class distresses through the use of convolutional neural network (CNN).

The organization for the remainder of this paper is detailed in the following sentences. Section 2 introduces the training data preparation including the data collection, crack image labelling process, and standards. Section 3 discusses the methodology and architectures of the proposed networks. Section 4 presents the process of model evaluation and the image classification testing. Section 5 concludes the research and discusses the results.

## 2. Data Preparation

In this paper, a novel approach based on a CNN architecture is proposed to achieve three different goals: crack image detection, pavement marking and crack image classification, and crack severity classification. The research consists of four stages, summarized as follows:

1) Establishment of pavement crack image dataset with different crack severity levels
2) Data preprocessing
3) Creation of multilayer CNN architecture
4) Model evaluation, including model accuracy and image class prediction.

To limit potential error, it was decided that the first generation of the proposed CNN should focus on pavements which exhibited fatigue cracking. Since CNNs fall under the umbrella of machine learning, it should be understood that one of the major goals in developing the architecture of the CNN is the limited mimicry of some level of human behavior. In this particular scenario, fatigue cracking is readily identifiable to the trained human eye, whether through direct observation or observation of an image of the distress. Accordingly, the architecture developed should be able to successfully mimic this behavior after some level of training.

### 2.1. Establishment of the Image Library

An image library was created to feed CNN network with a dataset of different examples of pavement cracking images for supervised learning. The image library used consists of more than 4,000 high-resolution pavement images with size of 2016x1512, corresponding to approximately 3,000,000 pixels per photograph. The photographs selected for the database contain diverse variations of both pavement fatigue cracks and fine pavement surface textures. The images in the library were fed through the CNN to achieve two distinct objectives: facilitation of the differentiation between non-cracked pavement surfaces and pavement surfaces exhibiting cracking, and facilitation of the differentiation between severity levels of pavement cracks. All the pavement images in the library were collected in the town of Blacksburg, Virginia, USA in the spring of 2019. Photographs were taken at various locations around the town, and the photographer's walking speed remained constant when photographing the distresses. Images photographed include non-cracked asphalt pavement surfaces, asphalt pavement surfaces with minor cracking, asphalt pavement surfaces with major cracking, and asphalt pavement surface images with markings or centerline. Images were taken using an iPhone 7 with a 12MP lens in sunny, cloudy and rainy conditions in order to consider natural variations in lighting corresponding to changes in the weather. Fig.1 shows the location of the taken images in the town of Blacksburg. Images were primarily gathered from residential areas due to safety considerations.

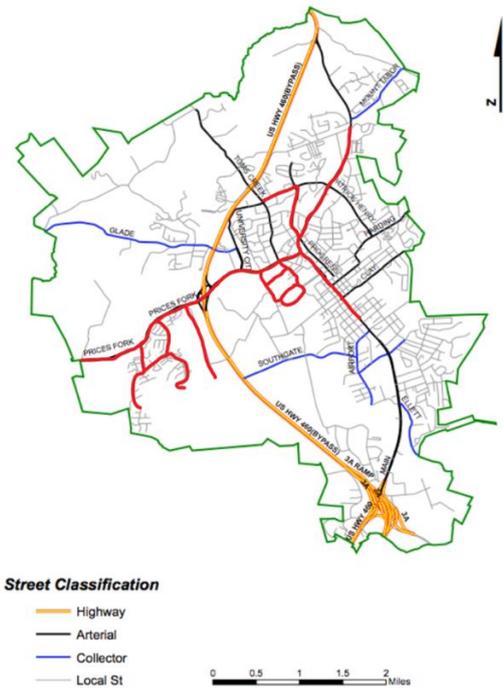

Fig. 1. The location of pavement surface images taken in the town of Blacksburg (red lines).

Upon completion, the image library contained images of all types of cracking with varying levels of severity as defined by the LTPP Distress Identification Manual (John S. Miller and William Y. Bellinger). The ground-truth of cracks on all images were classified with three round supervision by the data preparation team. An inspection by experts was conducted to ensure the accuracy of human classification and labeling. 2892 images from the library were chosen at random to be used in the training and subsequent testing of the model. More specifically, the training dataset consisted of 2024 images and the testing dataset consisted of 868 images.

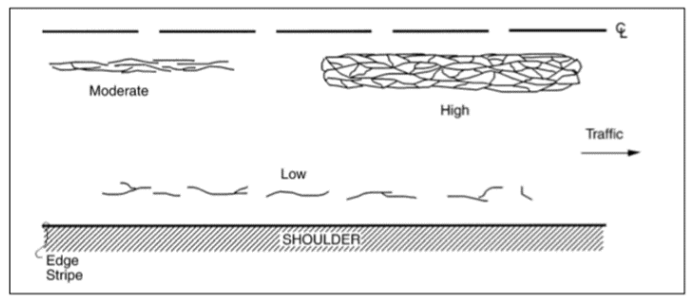

Fig. 2. Fatigue Cracking Severity Levels for Asphalt Concrete Pavements.

An additional 677 pavement marking images were collected to achieve the ability to distinguish pavement with markings from all other pavement images. Some of the images within the image library contained indications of both distresses and pavement markings.

Table 1

Comparison to Existing Databases.

| Name of Database | Channel Type | Number of Images | Image Resolution |
|---|---|---|---|
| AigleRN | Greyscale | 38 | 320x480 |
| CFD | RGB | 118 | 991×462/311×462 |
| Database(built) | RGB | 3569 | 2016x1512 |

*2.2. Establishment of the Crack Criteria*

The database created contains 3569 images, each of which contains many different features. For the crack detection experiment, 1592 images were assigned as training and testing data. The images were manually labeled into two classes. For the mark and crack classification, there were 677 images which were labeled into two classes. Moreover, 2892 images in the database were labeled into three classes including pavement without cracks, moderate severity cracks, and high severity cracks. The data preparation team followed the guidance defined by LTPP for fatigue cracking. According to the LTPP Distress Identification Manual (Fourth Revised Edition), severity levels for fatigue cracking can be defined as either low, moderate, or high. The main differences among these severity levels are the presence of connection between cracks, spalling of cracks, and presence of surface movement.

Following the Distress Identification Manual criteria, an image classification team was gathered to identify pavement distress existence, pavement distress levels, and presence of traffic marking images among the collected dataset. Fig. 2, provided in the DIM, provides an example visual illustration of the differences across fatigue cracking severity levels. Fig. 3 shows an example of smooth pavement and pavement with cracking, and Fig. 4 and Fig. 5 show the manually classified images compared with defined example in the Manual. In Fig. 4 and Fig. 5, the left-hand images were taken by the research team, and the right-hand images are the examples provided in the DIM. Approximately 3000 images were used for crack severity classifications, and approximately 1500 images were used for crack presence classification.

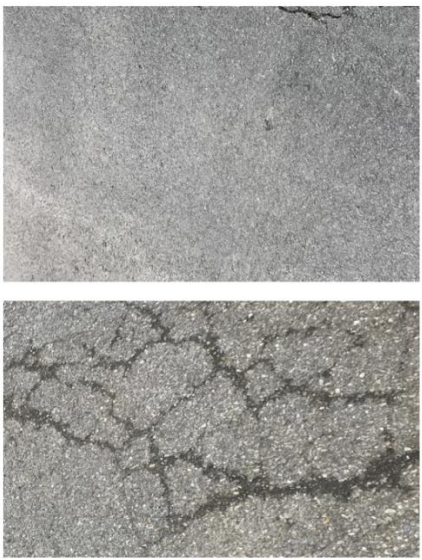

Fig. 3. Pavement surface without cracking (top) and with cracking (bottom).

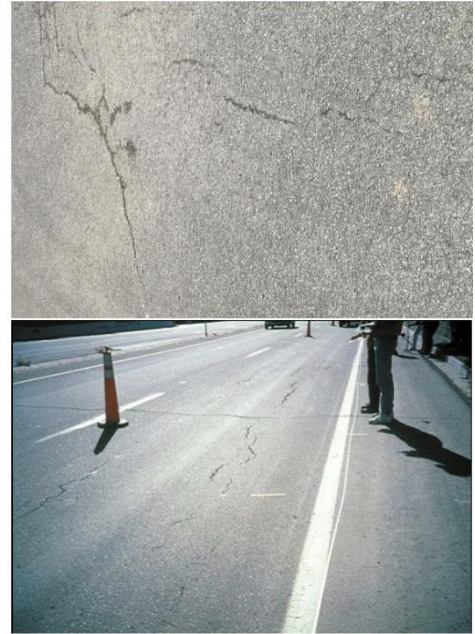

Fig. 4. Low Severity Fatigue Cracking.

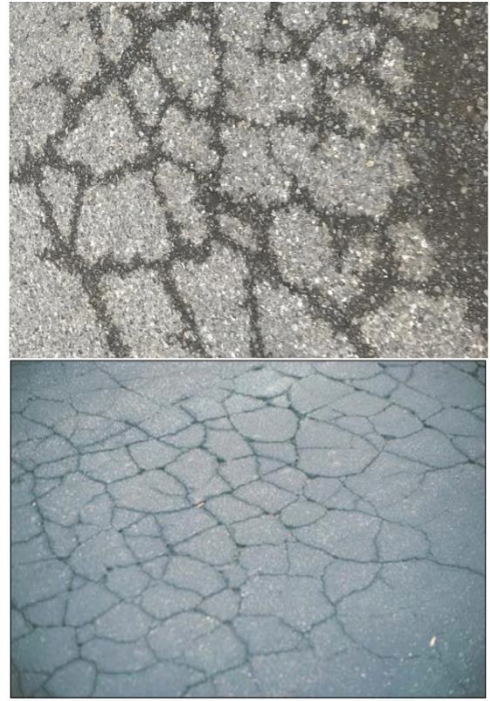

Fig. 5. High Severity Fatigue Cracking.

*2.3. Image Processing*

Since the size of the training image dataset was relatively large, there existed a potential for overload of the computer being used to process the data. To reduce the risk of computational overload, all training images were resized from their original resolution to 500x500 before feeding the dataset to the CNN model for training. Additionally, all images were converted to gray-value images from their original RGB state prior to being fed to the CNN. This image pre-processing significantly shortened the time and load of the CPU training process, while also retaining most of the information from the original images that contributes to the accuracy of the training.

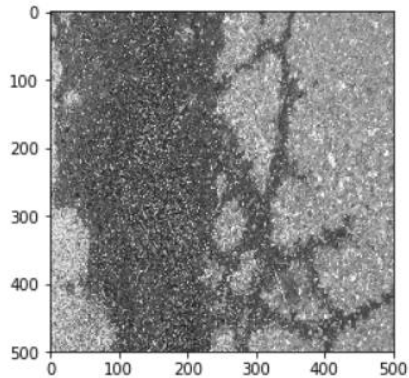

Fig. 6. An example of resized gray-value training image.

As Fig. 6 shows, the distress severity level information remains readily recognizable in the downsized training image. Since the computer identifies the presence of cracks by individual pixel shade, the gray-value image is sufficient for the computer to train and learn. After image preprocessing was completed, labeling was completed to ensure training accuracy.

## 3. Methodology

Deep learning is a subset of machine learning methods based on artificial neural networks. Among the many deep learning architectures, deep neural networks (DNNs), convolutional neural networks (CNNs), and recurrent neural networks (RNNs) are most frequently used. In deep learning, feature extraction does not rely on manual work, but rather on automatic extraction completed by a machine.

A convolutional network (LeCun, 1989) is a kind of neural network that is specially used to process data with known grid topology. Convolutional neural networks (CNN or ConvNet) are a class of deep, feedforward neural networks with local connection and weight sharing characteristics. A typical CNN structure usually consists of input layer (typically image dataset), a convolutional layer, a pooling layer, and a fully connected layer.

The convolution layer is a parallel feature maps set, which is created by sliding different convolution kernels on the input image and performing certain operations. Convolutional neural networks can effectively reduce the dimensionality of images with large amounts of data to achieve an image with relatively smaller amount of data (without affecting the result), and can retain the characteristics of images, similar in principle to the vision capabilities of human beings. Feature maps are extracted from the input data through the convolution layers which contain multiple convolution kernels. The elements of the convolution kernel correspondingly point to a weight coefficient and deviation. The parameters of the convolutional layer include the size of the kernels, the stride, and the padding. The parameters directly dictate the size of the output feature graph of the convolutional layer. When the convolution kernel size is larger, more complex features are extracted from final layers.

The main purpose of the convolutional layer is to retain the image features. The pooling layer serves the purpose of dimensional reduction of the data, which is an effective technique to avoid overfitting. The full connection layer is used to output the task-dependent desired results. The convolution stride defines the distance between the convolution kernels when they are scanning the feature graph adjacent to one other. When the stride is 1, the convolution kernels will scan the elements of the feature graph one by one. When the stride is n, n-1 pixels will be skipped in the next scan process.

Pooling is a reduction sampling method used in the CNN. The pooling layer has various non-linear forms. The space of the data from the convolutional layer needs to be reduced for decision-making. The pooling layer always follows the convolutional layer and continuously reduces the data space. This downsize controls overfitting to a certain extent.

For the crack detection function, the main target for the network is to automatically recognize the image of pavement fatigue cracks among all pavement surface images in a given dataset. The target for mark-crack classification is to see if the network can detect cracks in addition to the road mark noise. The images collected for crack detection and mark-crack have a relatively smaller size compared to other images collected. The inputs for these functions were downsized to 256x256 pixels to reduce the computational load.

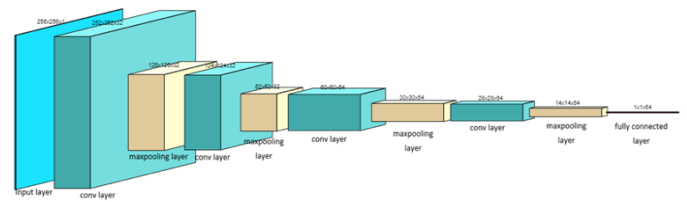

Fig. 7. Architecture of proposed CNN for crack detection and mark-crack classification.

As illustrated in Fig. 7, the CNN proposed in this study consists of four convolutional layers with ReLU and four, 2x2 max-pooling layers. The kernel size of four convolutional layers are 32x5x5, 32x3x3, 64x3x3, and 64x3x3, respectively. Two fully connected layers follow the convolutional layers. A cross-entropy loss function is also employed to calculate the performance of a training quality of a classification model.

For the crack severity classification function, the input size of 500x500x1 will keep the raw pixel values of the image, but the dimension (also called depth) should be set to one for grayscale images. The convolution layer receives the computation output of neurons linked with an input region. Each neuron calculates the dot product between the small region that is connected to the input volume and the weights. The ReLu layer then applies a basic activation function to keep the volume size the same. The FC (or fully connected) layer then calculates the class score.

As shown in Fig. 8, each convolutional layer is followed by a maxpooling layer to subsample the feature map. The first convolutional layer has 32 kernels, which corresponds to the depth of the convolutional layer and also to the amount of feature maps. The filter size of the first convolutional layer is 5x5, and the padding type is set as valid. The input image will be calculated and transferred to a feature map of size 496.

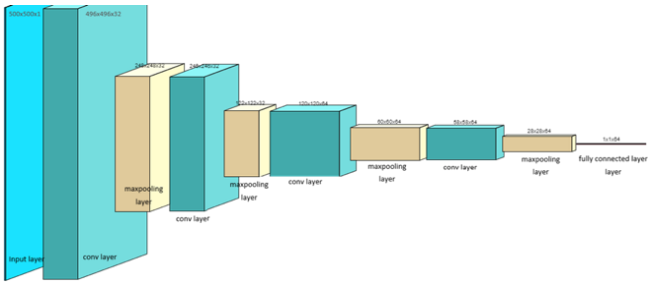

Fig. 8. Architecture of proposed CNN for crack detection and mark-crack classification.

The feature map size is calculated by the following output size shrinks by filter size – 1. Next, the output is passed through the maxpooling layer, the size of the maxpooling filter is 2x2 and the stride is 2. This process is for down sampling the feature map.

## 4. Results

### 4.1. Fatigue Crack Detection Test

In the training process, all experiments are performed using an Intel(R) Core i-7-7700HQ CPU@ 2.80GHz CPU with 16GB RAM and NVidia 1060 with 6GB RAM GPU. As the resolution of the training dataset used is relatively large, and the number of training images is also large, input images are resized to 256x256. A batch size of 64 is selected and the validation split function is applied to split the 1592 training images into training set and testing set with a ratio of 0.2. This results in 1273 training sample and 319 samples for validation. The output of the test is a binary Boolean-like result falling into one of two categories: non-cracked pavement surface or pavement surface with fatigue cracking present.

Fig. 9 shows the training results of the proposed CNN model performance on the crack detecting function. The objective of the training was to enable the model to classify the pavement surface images into two categories: non-cracked pavement surface or pavement surface with fatigue cracking present. The results show the loss of the training process is approaching zero while the accuracy of the classification model increased from 79.03% to 96.24% before and after training. As there may exist a small number of human errors during the labeling step, the accuracy is ideal for pavement surface fatigue cracking identification.

### 4.2. Crack and Marking Classification Test

For the testing of the crack and pavement mark classification function the input images are also resized to size of 256x256. The batch size is 64 and the validation split function is applied to split the 677 training images into training set and testing set with a ratio of 0.1. This resulted in 552 training images and 62 testing images. The output has two classes: pavement marking and cracked pavement with pavement marking.

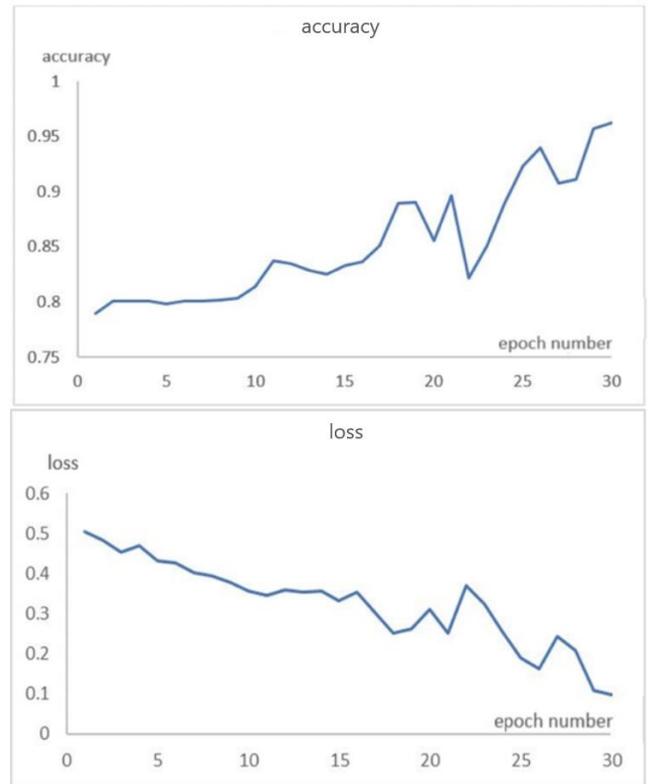

Fig. 9. Accuracy and loss on training data for crack detection.

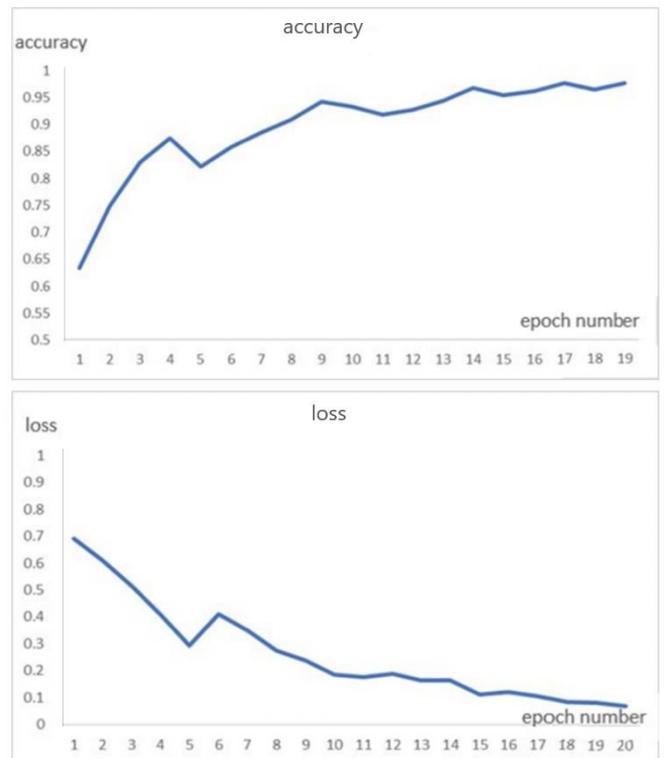

Fig.10. Accuracy on training data for mark/crack classification.

As Fig.10 shows, there are 20 epochs for the training process instead of 30 as initially developed. While the model was being trained, it was noticed that the accuracy trend began to decrease, and the loss trend began to increase after the 20[th] epoch. The method employed to prevent a decrease in accuracy is known early stopping. Early stopping is a method by which the performance of the model on the test set is monitored during training. The training process is then stopped at the moment that the performance of the model on the test set begins to decrease, so as to avoid the problem of over-fitting caused by continued training. Thus, the training of the model was stopped at the 20th epoch to avoid over-fitting error. To further avoid over-fitting problems, the training image dataset is enlarged at the early steps. The dataset built at the first stage contains 500 hundred images in total. The accuracy of the first attempts is around 0.3 and the validation accuracy is 0. The other effort to overcome over-fitting problem was adding cross-validation before training the model which allows the test set as a truly unseen dataset.

*4.3. Fatigue Cracking Severity Test*

The same architecture was used to test the severity level classification accuracy. For the testing of this function, the images are resized to 500x500. Despite this, the model still took three hours to train for 30 epochs. The batch size is selected as 64, and the validation split function is applied to split the 2892 training dataset into model training and testing datasets with a ratio of 0.3. This resulted in a training set of 2024 samples and validation set of 868 sample images. The training dataset consists of three categories: pavement surfaces with no cracking, pavement surfaces with moderate cracking, and pavement surfaces with high-severity cracking. During data preparation, the three-category training dataset is labeled as a category 0, 1, or 2, representing the respective levels of cracking severity.

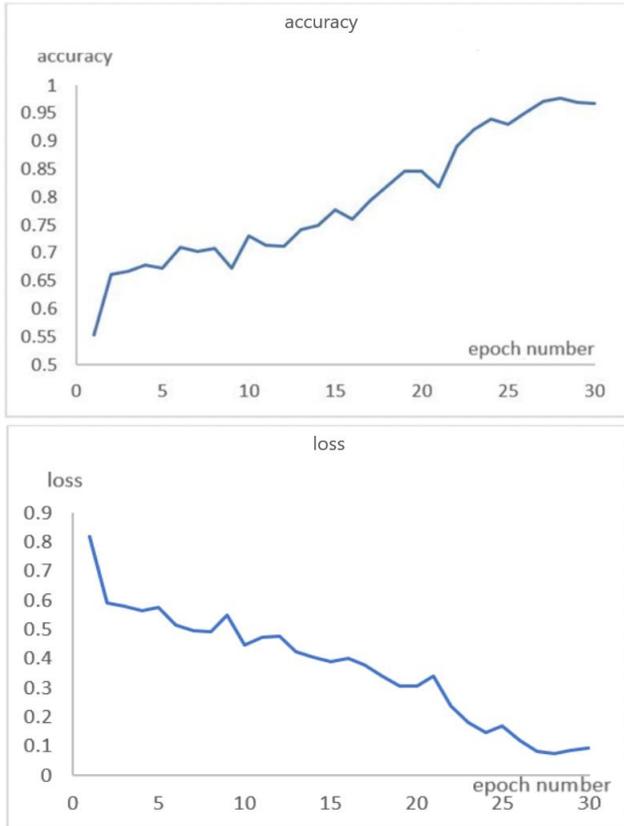

Fig. 11. Accuracy on training data for crack severity classification.

Fig. 11 shows the training results of the proposed CNN model when trained for the crack severity classification function. The results show the loss of the training process is approaching zero while the accuracy of the classification model increased from 55% (which is almost random) to 96.74%. Considering the reality that there may exist some level of human error during the labeling step, the accuracy is ideal for pavement surface fatigue cracking severity level classification.

*4.4. Comparison to Existing CNN Crack Detection Methods*

Automated crack detection is a trending topic in the civil engineering, infrastructure management, and computer science research areas. Existing methods are using CNN technologies to detect crack pixels and determine the morphology of detected cracks. Table 2 shows a comparison of the five existing approaches including the deep learning-based method [24] based on a VGG-net.

Table 2
Crack pixel detection accuracy comparison on CFD dataset.

| Method | Precision (%) |
|---|---|
| Zou et al. (2012) | 73.22 |
| Avila et al. (2014) | 78.56 |
| Ronneberger et al. (2015) | 92.15 |
| Li et al. (2018) | 89.90 |
| Nguyen et al. (2018) | 95.67 |

Cubero-Fernandez et al [25] presented a decision tree algorithm to complete the image classification in 2017. They used 400 images including transverse cracking, longitudinal cracking, non-cracking, and fatigue cracking. The work achieved an 80% accuracy in detecting the crack type. As a comparison, the model presented here has an accuracy of 96.23% for crack and non-crack classification, a 97.64% accuracy for mark and crack classification, and a 96.74% accuracy for crack severity level classification.

It should be noted that research for classification and decision-making for pavement fatigue cracking, marking. and crack severity is very limited. One reason that this is the case is due to the limited size of the existing databases. The amount of data required for other image recognition tasks that have proven successful, such as facial recognition models, well exceeds the size of the existing databases for pavement crack recognition. Accordingly, to achieve a successful model, a much larger data set was required. The database established in this study has a much larger size and better resolution when compared with the existing databases. This is a key component that is required to achieve the level of accuracy reached by the proposed model. Additionally noteworthy is the fact that the crack severity criteria image classification is one of the first attempts at such a classification in this field. Beyond remedying database size issues, the architecture created was intentionally designed to be of limited complexity. The relatively simplistic approach of the proposed algorithm prevented the need for additional complex computation, thus increasing computational efficiency, and reducing the required training time. Accordingly, the method proposed has good potential for real-world application in automated pavement crack detection and pavement rating systems.

*4.5. Pavement Crack Image Prediction*

After training the model for three different functions, the trained CNN was successfully able to classify cracked and non-cracked pavement surfaces, classify the crack severity level of test images, and differentiate between pavement images which contained only markings and those which contained both markings and cracks.

ModelC, ModelM and ModelS are pretrained models of the proposed CNN. Test images, labeled "test_n" (where n is the number of the test) were prepared to include images indicative of non-cracked pavement surfaces, pavement surfaces with different severity levels of fatigue cracking, and pavement surfaces with mark and with or without alligator cracking. In the prediction phase, 50 images of each category including crack, non-crack, mark, moderate severity crack, high severity crack were prepared for the proposed model to make predictions.

Fig. 12 illustrates examples of the results of the proposed model's predictions of crack presence in prepared test images (test_4, test_9, test_6 and test_8). The prediction results correctly match crack presence classification completed by humans. The prediction accuracies of crack and non-crack images are 46 out of 50 and 48 out of 50, respectively. The overall accuracy of crack image classification from the test dataset is 94%.

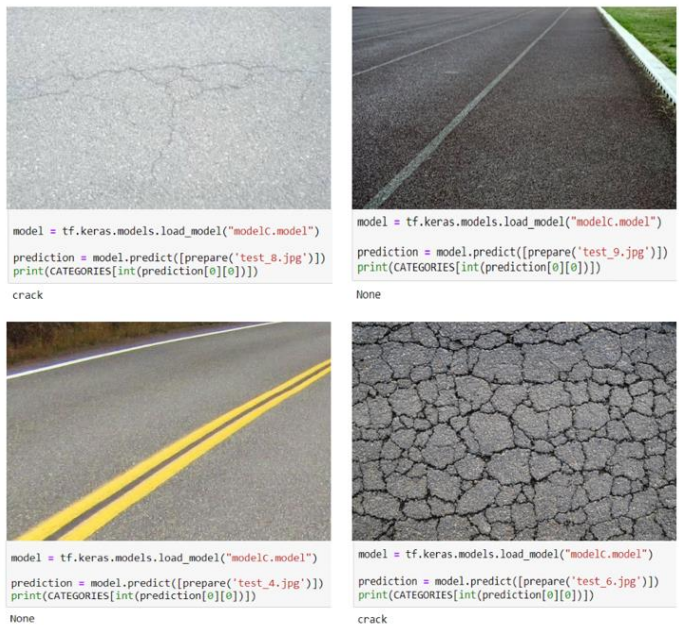

Fig. 12. Crack and Non-Crack Image Classification.

The second test that the proposed model completed was for the classification of the presence of pavement marks and cracking. The purpose of this classification test was to assess whether the presence of pavement marks affect the model's ability to recognize cracking. Fig.13 shows the classification results of test_1, test_4, test_10 and test_11. The prediction accuracy of mark and crack classification is 49 out 50 and 46 out of 50, respectively. The overall prediction accuracy is 95%.

Fig. 14 shows the classification of different crack severity levels. The output categories are no cracking, moderate severity cracking, and high severity cracking, and in the code, these categories are represented as none, minor, and major, respectively. The prediction accuracy of non-crack, moderate severity crack and high severity crack is 48 out of 50, 42 out of 50, and 46 out of 50 respectively. The overall prediction accuracy of crack severity is 90.67%. Table 3 provides a summary of training and prediction accuracy of the model when trained with different sets of image data.

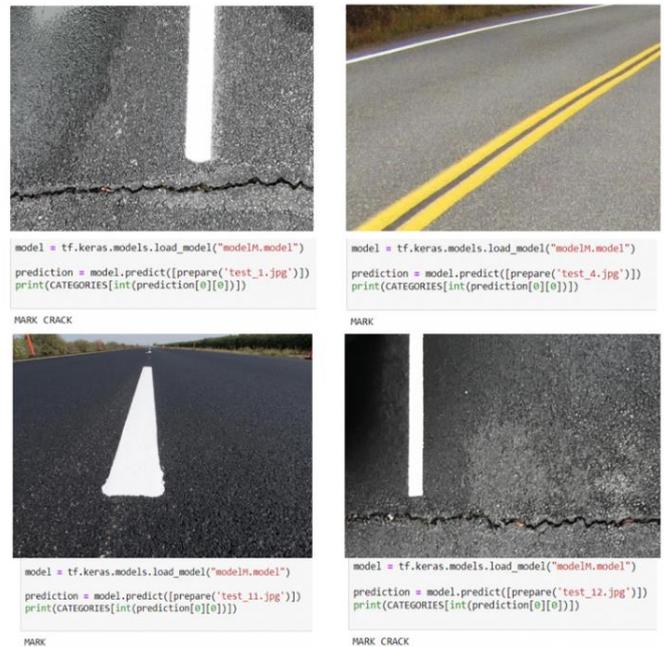

Fig. 13. Mark and Crack Classification Results.

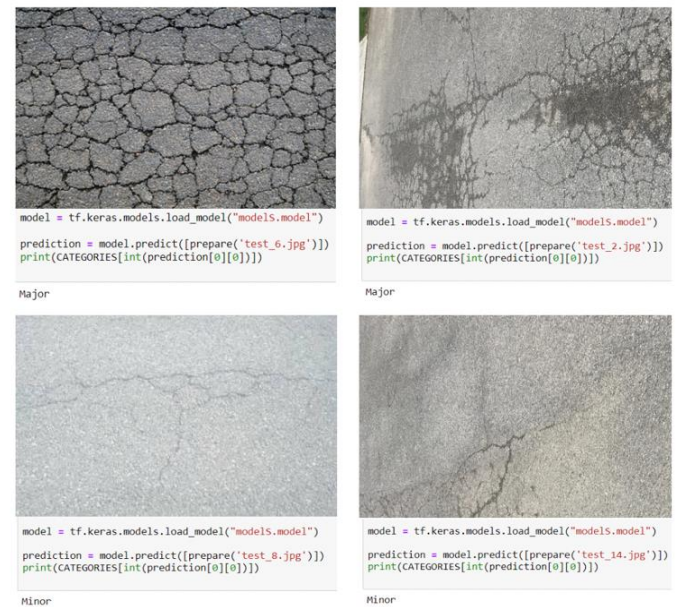

Fig. 14. Examples of Crack Severity Classification Results.

Table 3
Comparison between training and prediction accuracy of different crack categories.

|  | Crack/Non-Crack | Mark/Crack | Crack Severity |
|---|---|---|---|
| Model name | C | M | S |
| Training accuracy | 96.23% | 97.64% | 96.74% |
| Prediction accuracy | 94.00% | 95.00% | 90.67% |

## 5. Conclusion and Future Research

In this project, a Convolutional Neural Network (CNN) was proposed to classify pavement surface images with the following features: crack and non-crack, mark and crack, and crack severity level. A well-labeled, high resolution pavement fatigue cracking image bank was established. The dataset includes over 4000 images collected by iPhone 7 in the town of Blacksburg. The data processing team were trained and used two-round inspection under the guidance of Distress Identification Manual by FHWA LTPP project to manually identify, categorize, and label distress images.

The proposed 4-layer CNN with Relu and maxpooling layers achieved the objective of crack detection and severity level classification with high resolution input data. The experimental results show that the proposed model performs well when supplied with high quality input image data. The accuracy of the trained model for classification of cracked and non-cracked pavement surface images is 96.23% after 30 epochs. For marking and crack classification, the accuracy reaches 97.64% after 20 epochs training. The accuracy of the model for classification of cracking severity level of no cracking, moderate severity cracking, and high severity cracking achieved 96.74% after 30 epochs. The training process took one hour for both the crack existence classification and mark crack classification, and three hours for the crack severity level classification training. The training time is relatively long due to the high resolution of resized image and the complexity of the established model. The CNN model proposed outperformed the existing models for the aforementioned classification tasks.

As a part of pavement management system, roadway pavement surveys can be time consuming and costly, and manual pavement image processing is highly labor intensive. The deep learning method proposed has the potential to work well in pavement management systems by identifying images that contain fatigue cracking from within a massive databank of pavement images, eliminating the need for manual processing. Additionally, the severity level classification function could be a handy tool for pavement condition surveying and pavement condition and performance ratings that will be used for pavement management system.

While the research presented is primarily focused on the classification of fatigue cracking in asphalt pavements, the deep CNN model proposed has a positive outlook for use in other pavement distress-related feature extraction scenarios. The ubiquity of distresses on roadway pavements provides an excellent opportunity to refine the proposed model to automate the distress classification process. In future research, we will seek to establish different versions of our CNN architecture to adapt various types of other pavement distresses, such as longitudinal cracking, edge cracking, potholes, and other pavement distresses. We will then seek to create a synchronous system architecture that is capable of identifying all visible distresses on the pavement. Furthermore, we will refine the distress severity level classification process, creating a more concise and detailed model that is capable of determining overall pavement condition in a single test. We will also seek to include geographical information in the image databank to facilitate integration with GIS-based pavement management systems. Our ultimate vision is to produce an automated system capable of completing comprehensive pavement surveys by determining composite distress level of pavement sections. Such a system would also be able to provide a quantitative rating for this composite section, and aid in the decision-making and pavement management processes.